\documentclass{esannV2p}
\usepackage{graphicx}
\usepackage[latin1]{inputenc}
\usepackage{amssymb,amsmath,array}
\usepackage{booktabs}
\usepackage{multirow}
\usepackage{xcolor}
\usepackage[square,numbers]{natbib}
\usepackage{enumitem}
\usepackage{xfrac}
\usepackage{watermark}
\watermark{\thispageheading{\sffamily Proceedings of the 30th European Symposium on Artificial Neural Networks, Computational Intelligence and Machine Learning (ESANN 2022), pp. 491-496 doi:10.14428/esann/2022.ES2022-58}}
\definecolor{bestresult}{gray}{0.9}
\newcommand{\best}[1]{\colorbox{bestresult}{#1}}
\begin{document}
\title{Beyond Homophily with\\ Graph Echo State Networks}

\author{Domenico Tortorella and Alessio Micheli
%
%
\vspace{.3cm}\\
%
University of Pisa - Department of Computer Science \\
Largo B. Pontecorvo 3, 56127 Pisa - Italy
%
}

\maketitle

\begin{abstract}
Graph Echo State Networks (GESN) have already demonstrated their efficacy and efficiency in graph classification tasks.
However, semi-supervised node classification brought out the problem of over-smoothing in end-to-end trained deep models, which causes a bias towards high homophily graphs.
We evaluate for the first time GESN on node classification tasks with different degrees of homophily, analyzing also the impact of the reservoir radius.
Our experiments show that reservoir models are able to achieve better or comparable accuracy with respect to fully trained deep models that implement ad hoc variations in the architectural bias, with a gain in terms of efficiency.
\end{abstract}

\section{Introduction}
Graphs provide a useful structure to represent relations between entities, such as paper citations or web page networks.
A plethora of neural models have been proposed to solve graph-, edge-, and node-level tasks \cite{Bacciu2020}, most of them sharing an architecture structured in layers that perform local aggregations of node features.
This architectural bias, where node features are progressively smoothed in deeper layers via local aggregation \cite{Li2018}, is the source of most of the issues that graph neural models are facing.
This bias towards locally homogeneous graphs is more apparent in node classification tasks, where graphs presenting a significant number of inter-class edges, i.e. a low \emph{homophily} degree, present a challenge to convolutive models.
Graph Echo State Network (GESN) \cite{Gallicchio2010} is an efficient model within the reservoir computing (RC) paradigm.
In RC, input data is encoded via a randomly-initialized reservoir, while only a linear readout requires training.
GESN has already been successfully applied to graph-level classification tasks \cite{Gallicchio2020}.
In this paper, we analyze for the first time its application to node classification tasks, focusing in particular on the efficacy in tackling low-homophily graphs.

\section{Node classification and homophily}\label{sec:background}
Let $\mathcal{G} = (\mathcal{V}, \mathcal{E})$ denote a graph with node feature vectors $\mathbf{u}_v \in \mathbb{R}^U$ for each node $v \in \mathcal{V}$.
We also denote by $\mathcal{N}_r(v)$ the set of nodes within $r$ hops of node $v$, and by $\mathbf{A}$ the graph adjacency matrix.
The goal of a semi-supervised node classification task is to learn a model from a subset of graph nodes with known target labels $\{(\mathbf{u}_v, y_v)\}_{v \in \mathcal{V}_{\mathrm{train}}}$, in order to infer the node labels $y_v \in \{1, ..., C\}$ for the remaining nodes $\mathcal{V} \setminus \mathcal{V}_{\mathrm{train}}$ using the network structure and input features $\mathbf{u}_v$.
Most common graph convolutional models are structured in $L$ layers, where each layer learns an embedding for each node based on an increasingly large receptive field.
These layers can be formalized as \cite{Xu2019}
\begin{equation}\label{eq:gnn-layer}
  \mathbf{h}_v^{(\ell)} = \textsc{combine}\left(\mathbf{h}_v^{(\ell-1)}, \textsc{aggregate}(\{\mathbf{h}_{v'}^{(\ell-1)} : v' \in \mathcal{N}_1(v)\})\right),
\end{equation}
where node embeddings $\mathbf{h}_v^{(\ell)} \in \mathbb{R}^H$ of layer $\ell$ are obtained by aggregating the previous embeddings $\mathbf{h}_{v'}^{(\ell-1)}$ of node $v$'s $1$-hop neighbors via $\textsc{aggregate}(\cdot)$, and then combined with the node's previous embeddings $\mathbf{h}_v^{(\ell-1)}$ via $\textsc{combine}(\cdot)$; for $\ell = 1$, $\mathbf{h}_v^{(0)} = \mathbf{u}_v$.
The final layer $L$ either directly predicts the one-hot encoding of target label $y_v$, or is followed by an MLP that serves this purpose.
The whole model is trained end-to-end by typically minimizing the cross-entropy loss.

The choice of functions in \eqref{eq:gnn-layer} determines the architectural bias of the model.
For example, GCN \cite{Kipf2017} layers are defined as $\mathbf{h}^{(\ell)} = \mathrm{relu}(\mathbf{\hat{A}} \mathbf{h}^{(\ell-1)} \mathbf{\Theta}^{(\ell)})$,
where $\mathbf{\hat{A}}$ is the normalized adjacency matrix, $\mathbf{\Theta}^{(\ell)}$ are learnable weights, and $\mathbf{h}^{(\ell)}$ is the row stack of node features for layer $\ell$.
It has been shown that stacking more than three or four layers of graph convolution causes a degradation in accuracy \cite{Li2018}, since representations $\mathbf{h}_v^{(\ell)}$ converge asymptotically to a fixed point of $\mathbf{\hat{A}}$ as $\ell$ increases, or more generally, to a low-frequency subspace of the graph spectrum.
This problem is known as \emph{oversmoothing}.
Indeed, by acting as a low-pass filter, GCNs are biased in favor of tasks whose graphs present a high degree of homophily, that is nodes in the same neighborhood mostly share the same class \citep{Zhu2020}.
Formally, homophily in a graph can be quantified \citep{Zhu2020} as the intra-class edges ratio
\begin{equation}\label{eq:homophily}
  \mathfrak{h}_\mathcal{G} = \left|\{(v, v') \in \mathcal{E} : y_v = y_{v'}\}\right| / \left|\mathcal{E}\right|.
\end{equation}
Changes in the model architectural bias have been proposed to improve classification on low homophily graphs.
Some solutions individuated by \cite{Zhu2020} are:
\begin{enumerate}[itemsep=0mm]
  \item separate ego and neighborhood representations in \eqref{eq:gnn-layer}, by aggregating on open node neighborhoods $\mathcal{N}_r(v) \setminus \{v\}$ and combining by concatenation;
  \item extend aggregation to multi-hop neighborhoods $\mathcal{N}_r(v)$, $r > 1$, e.g. as in graph convolutions with Chebyshev polynomial filters \cite{Defferrard2016};
  \item exploit also the representations $\mathbf{h}_v^{(\ell)}$ computed at each intermediate layer $\ell < L$ to make predictions, e.g. as in Jumping-Knowledge networks \cite{Xu2018}.
\end{enumerate}
H2GCN \cite{Zhu2020} incorporates all three architectural solutions.
Alternative solutions include altering the graph structure to improve the homophily degree, in order to increase the ratio of intra-class edges in node neighborhoods \cite{Gasteiger2019,Topping2022}.

\section{Reservoir computing for graphs}\label{sec:graphesn}
Reservoir computing is a paradigm for the efficient design of recurrent neural networks.
Input data is encoded by a randomly initialized reservoir, while only the task prediction layer requires training.
Graph Echo State Networks (GESNs) extended the reservoir computing paradigm to graph-structured data \cite{Gallicchio2010}, and have already demonstrated their effectiveness in graph classification tasks \cite{Gallicchio2020}.
Node embeddings are recursively computed by the dynamical system
\begin{equation}\label{eq:graphesn}\textstyle
  \mathbf{x}_v^{(k)} = \tanh\left(\mathbf{W}_{\mathrm{in}}\, \mathbf{u}_v + \sum_{v' \in \mathcal{N}_1(v)} \mathbf{\hat{W}}\, \mathbf{x}_{v'}^{(k-1)}\right),\quad \mathbf{x}_v^{(0)} = \mathbf{0},
\end{equation}
where $\mathbf{W}_{\mathrm{in}} \in \mathbb{R}^{H \times U}$ and $\mathbf{\hat{W}} \in \mathbb{R}^{H \times H}$ are the input-to-reservoir and the recurrent weights, respectively (input bias is omitted).
Equation \eqref{eq:graphesn} is iterated over $k$ until the system state converges to fixed point $\mathbf{x}_v^{(\infty)}$, which is used as the embedding.
The existence of a fixed point is guaranteed by the Graph Embedding Stability (GES) property \cite{Gallicchio2020}, which also guarantees independence from the system's initial state $\mathbf{x}_v^{(0)}$.
A necessary condition \cite{Tortorella2022} for the GES property is $\rho(\mathbf{\hat{W}}) < 1 / \alpha$, where $\rho(\cdot)$ denotes the spectral radius of a matrix, i.e. its largest absolute eigenvalue, and $\alpha = \rho(\mathbf{A})$ is the graph spectral radius.
This condition also provides the best estimate of the system bifurcation point, i.e. the threshold beyond which \eqref{eq:graphesn} becomes asymptotically unstable.
Reservoir weights are randomly initialized from a uniform distribution in $[-1,1]$, and then rescaled to the desired input scaling and reservoir spectral radius, without requiring any training.
While in graph-level task node features are aggregated to provide global embeddings, for node classification tasks we directly apply a linear readout to node embeddings $\mathbf{y}_v = \mathbf{W}_\mathrm{out}\, \mathbf{x}_v^{(\infty)} + \mathbf{b}_\mathrm{out}$, where the weights $\mathbf{W}_\mathrm{out} \in \mathbb{R}^{C \times H}, \mathbf{b}_\mathrm{out} \in \mathbb{R}^C$ are trained by ridge regression on one-hot encodings of target classes $y_v$.

The contractivity of \eqref{eq:graphesn} is a sufficient condition for the GES property \cite{Tortorella2022}.
However, the contractivity of graph convolution layers has also been linked to the degradation of representativeness in deep models \cite{Topping2022}.
Graph rewiring solutions to the homophily bias, such as \cite{Gasteiger2019}, greatly increase the edges of a graph, which in turn leads to an increase of $\alpha$ and a decrease in contractivity.
Therefore, in our experiments we will explore also values of the reservoir radius beyond the stability threshold, in this case by arbitrarily fixing the number of iterations of \eqref{eq:graphesn} to $K$.
Indeed, we can interpret the $K$ iterations of \eqref{eq:graphesn} as equivalent to $K$ graph convolution layers with weights shared among layers and input skip connections.
While in deep GCNs convergence to a fixed point of the graph convolution operator, due to stacking too many layers, has been linked to the oversmoothing issue \cite{Li2018}, GESNs can in principle avoid that by selecting a reservoir radius $\rho \gg 1/\alpha$.

\section{Experiments and discussion}\label{sec:experiments}

\begin{table}[h!]
  \footnotesize
  \centering
  \begin{tabular}{@{}l@{}c@{\hspace{-1pt}}c@{}c@{}c@{}c@{\hspace{-1pt}}c@{}c@{}c@{}c@{}}
    \toprule
    & \textsf{Texas} & \textsf{Wisconsin} & \textsf{Actor} & \textsf{Squirrel} & \textsf{Chameleon} & \textsf{Cornell} & \textsf{Citeseer} & \textsf{Pubmed} & \textsf{Cora} \\
    \midrule
    \textbf{Homo.} & $0.11$ & $0.21$ & $0.22$ & $0.22$ & $0.23$ & $0.30$ & $0.74$ & $0.80$ & $0.81$ \\
    \textbf{Nodes} & $183$ & $251$ & $7\mathord{,}600$ & $5\mathord{,}201$ & $2\mathord{,}277$ & $183$ & $3\mathord{,}327$ & $19\mathord{,}717$ & $2\mathord{,}708$ \\
    \textbf{Edges} & $295$ & $466$ & $26\mathord{,}752$ & $198\mathord{,}493$ & $31\mathord{,}421$ & $280$ & $9\mathord{,}104$ & $88\mathord{,}648$ & $10\mathord{,}556$ \\
    \textbf{Radius} & $2.56$ & $2.88$ & $9.99$ & $138.60$ & $61.90$ & $2.68$ & $13.74$ & $23.24$ & $14.39$ \\
    \textbf{Featur.} & $1\mathord{,}703$ & $1\mathord{,}703$ & $932$ & $2\mathord{,}089$ & $2\mathord{,}089$ & $1\mathord{,}703$ & $3\mathord{,}703$ & $500$ & $1\mathord{,}433$ \\
    \textbf{Classes} & $5$ & $5$ & $5$ & $5$ & $5$ & $5$ & $6$ & $3$  & $7$\\
    \midrule
    GCN & $59.5_{\pm 5.3}$ & $59.8_{\pm 7.0}$ & $30.3_{\pm 0.8}$ & $36.9_{\pm 1.3}$ & $59.8_{\pm 2.6}$ & $57.0_{\pm 4.7}$ & \best{$76.7_{\pm 1.6}$} & $87.4_{\pm 0.7}$ & \best{$87.3_{\pm 1.3}$} \\
    +JK & $66.5_{\pm 6.6}$ & $74.3_{\pm 6.4}$ & $34.2_{\pm 0.9}$ & $40.5_{\pm 1.6}$ & $63.4_{\pm 2.0}$ & $64.6_{\pm 8.7}$ & $74.5_{\pm 1.8}$ & $88.4_{\pm 0.5}$ & $85.8_{\pm 0.9}$ \\
    +Cheby & $77.3_{\pm 4.1}$ & $79.4_{\pm 4.5}$ & $34.1_{\pm 1.1}$ & $43.9_{\pm 1.6}$ & $55.2_{\pm 2.8}$ & $74.3_{\pm 7.5}$ & \best{$75.8_{\pm 1.5}$} & $88.7_{\pm 0.6}$ & \best{$86.8_{\pm 1.0}$} \\
    H2GCN & \best{$84.9_{\pm 6.8}$} & \best{$86.7_{\pm 4.7}$} & \best{$35.9_{\pm 1.0}$} & $36.4_{\pm 1.9}$ & $57.1_{\pm 1.6}$ & \best{$82.2_{\pm 4.8}$} & \best{$77.1_{\pm 1.6}$} & \best{$89.4_{\pm 0.3}$} & \best{$86.9_{\pm 1.4}$} \\
    \midrule
    MLP & \best{$81.9_{\pm 4.8}$} & \best{$85.3_{\pm 3.6}$} & \best{$35.8_{\pm 1.0}$} & $29.7_{\pm 1.8}$ & $46.4_{\pm 2.5}$ & \best{$81.1_{\pm 6.4}$} & $72.4_{\pm 2.2}$ & $86.7_{\pm 0.4}$ & $74.8_{\pm 2.2}$ \\
    \midrule
    GESN & \best{$84.3_{\pm 4.4}$} & \best{$83.3_{\pm 3.8}$} & $34.5_{\pm 0.8}$ & \best{$71.2_{\pm 1.5}$} & \best{$76.2_{\pm 1.2}$} & \best{$81.1_{\pm 6.0}$} & $74.5_{\pm 2.1}$ & \best{$89.2_{\pm 0.3}$} & \best{$86.0_{\pm 1.0}$} \\
    \bottomrule
  \end{tabular}
  \caption{Node classification accuracy on low and high homophily graphs (average and standard deviation; results of fully trained models reported from \cite{Zhu2020}; results within one standard deviation of the best accuracy are highlighted).}
  \label{tab:experiments}
  \vspace{-12pt}
\end{table}

\begin{figure}
  \centering
  \begin{minipage}{0.4\linewidth}
    \centering
    \includegraphics[scale=.7, trim=0cm 0.6cm 0cm 0.6cm, clip]{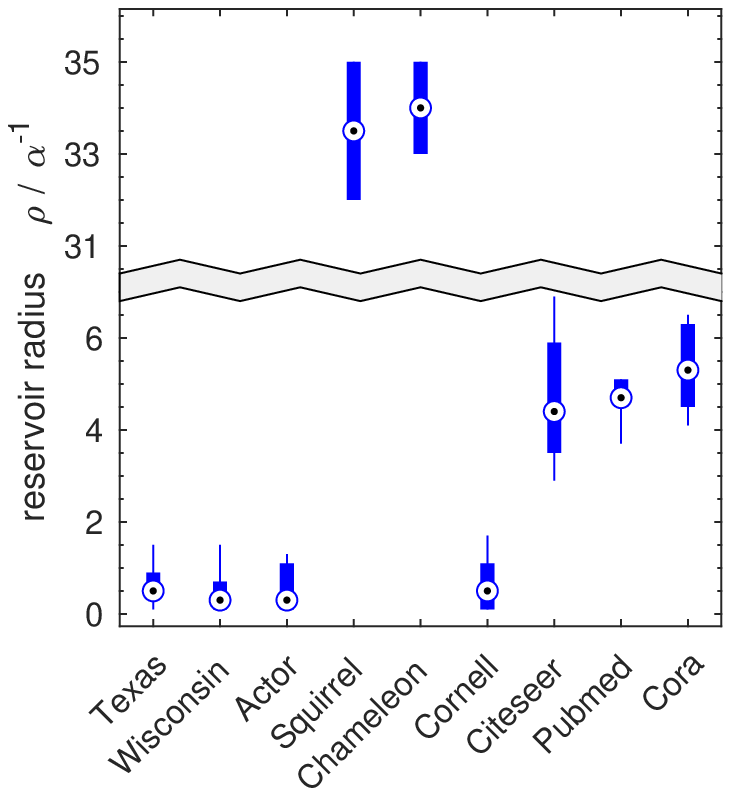}
    \caption{Reservoir radii selected on each task.}
    \label{fig:radii}
  \end{minipage}
  \hfill
  \begin{minipage}{0.55\linewidth}
    \centering
    \includegraphics[scale=.69]{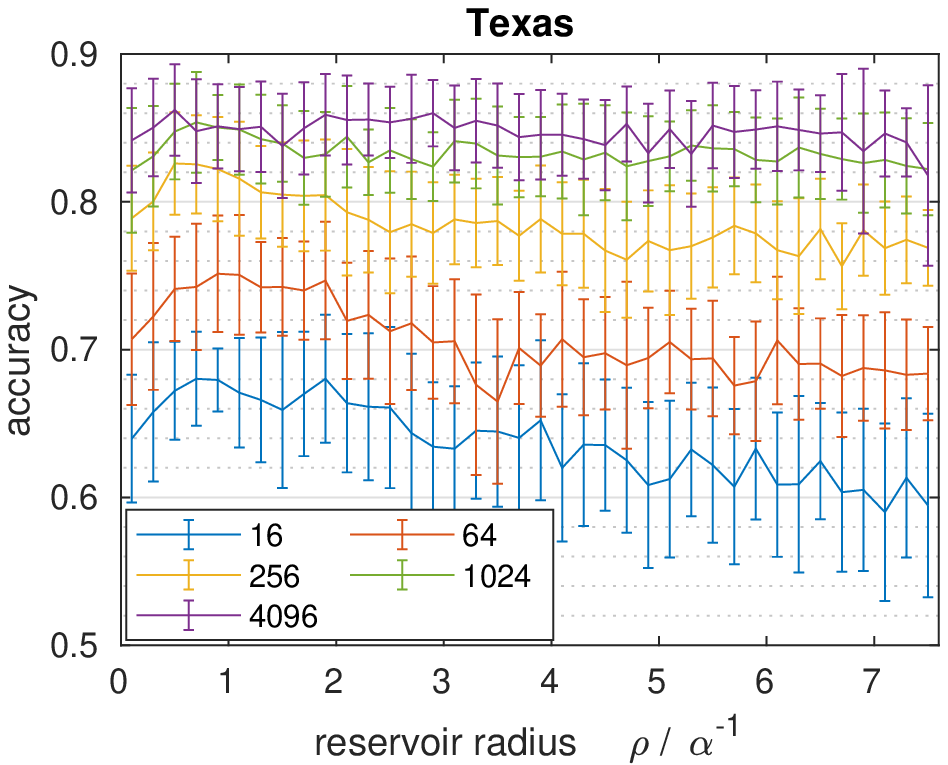}
    \caption{Impact of reservoir radius and units on classification accuracy for Texas.}
    \label{fig:texas}
  \end{minipage}
\end{figure}

\begin{figure}
  \centering
  \includegraphics[scale=0.7, trim=1.6cm 0.4cm 1cm 0.3cm, clip]{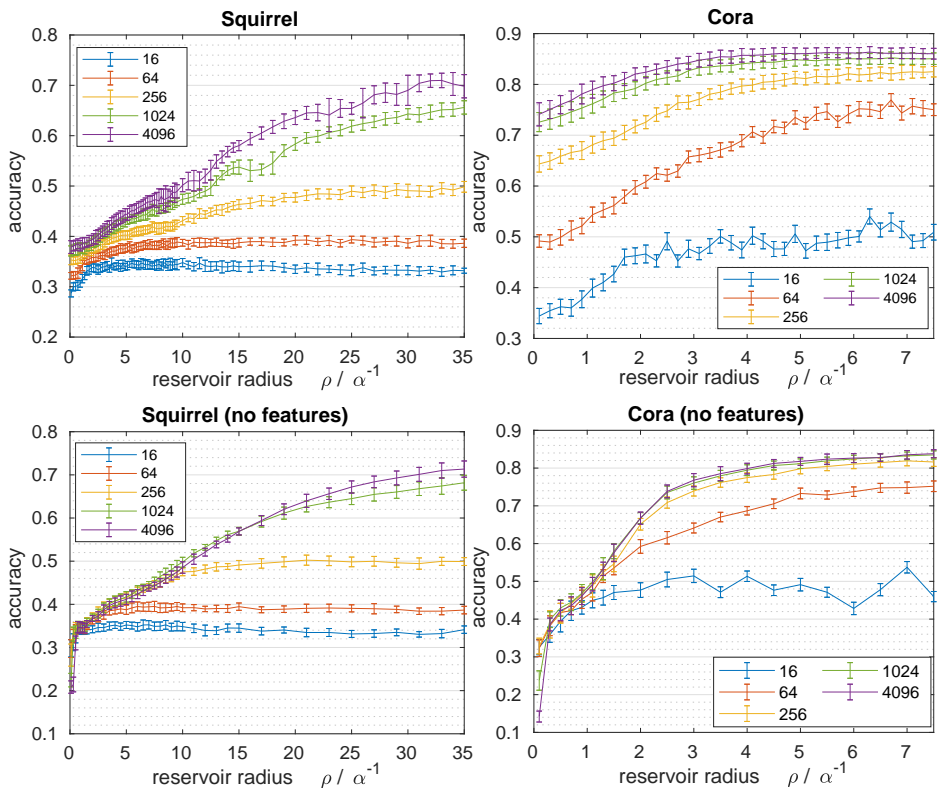}
  \caption{Impact of reservoir radius and units on classification accuracy for Squirrel ($\mathfrak{h}_\mathcal{G} = 0.22$) and Cora ($\mathfrak{h}_\mathcal{G} = 0.81$), with and without input features.}
  \label{fig:experiments}
\end{figure}

We evaluate GESN on six node classification tasks with low homophily degree ($\leq 0.3$) and three tasks with high homophily degree ($> 0.7$).
We adopt the same $10$ scaffold splits 48\%/32\%/20\% of \cite{Zhu2020}, averaging results in each fold over $10$ different reservoir initializations.
We explore a number of units ranging from $2^4$ to $2^{12}$, input scaling factors from $1$ to $\frac{1}{320}$, readout regularization values from $10^{-5}$ to $10^2$, and reservoir radii $\rho \alpha \in [0.1, 9.5]$ with steps of $0.2$ (up to $35$ with larger steps for Squirrel and Chameleon).
Embeddings are computed with at most $K = 100$ iterations of equation \eqref{eq:graphesn}.

Accuracy results are reported in Table \ref{tab:experiments}, while Fig. \ref{fig:radii} shows the reservoir radii selected in the $10$ splits.
We can observe three different behaviors, exemplified in Fig. \ref{fig:texas} and  \ref{fig:experiments} (top).
The number of reservoir units plays a significant role, offering best results when it is closer to the number of input features.
For Texas, Wisconsin, Actor, and Cornell, the performances of GESN are closer to the accuracies of MLP, which uses only node features $\mathbf{u}_v$, and H2GCN, with reservoir radii $\rho \alpha < 1$: in this case, the graph connectivity appears to be of no use.
While Squirrel and Chameleon present a low homophily degree, graph convolution models fare better than MLP: in this case graph connectivity needs to be taken into account.
On these two tasks, GESN improves upon the best model accuracy by $27.3\%$ and $12.8\%$, respectively, with reservoir radii selected in the range $33$--$35$.
Finally, on high homophily tasks (Citeseer, Pubmed, Cora) GESN performs generally in line with graph convolution models, which in turn do better than MLP; reservoir radii are selected in the range $4$--$6$.

We observe how the best accuracy results are for reservoir radii well above the stability threshold, which are required when the graph connectivity needs to be leveraged in classifying nodes.
To support our conclusion, in Fig. \ref{fig:experiments} (bottom) we report the accuracy on Squirrel and Cora where input features have been removed.
We observe that for stable embeddings ($\rho \alpha < 1$), accuracy significantly drops below the level reached by having input features, while it reaches almost the same levels of accuracy for the values of $\rho \alpha$ selected with features, which are well beyond the region where GESN stability is guaranteed.

Finally, we underline the efficiency of GESN.
Only the linear readout's $C (H+1)$ parameters require training, against the additional $O(H^2 L)$ parameters of models that need to be trained end-to-end through many gradient descent epochs (for further time comparisons, see \cite{Gallicchio2020}).
The time required to compute node embeddings and train the readout for a model of $4096$ units takes from $0.87$ to $1.58$ seconds on a GPU Nvidia Tesla V100, depending on graph size.

\section{Conclusion}\label{sec:conclusion}
For the first time, we have applied Graph Echo State Networks to the task of node classification.
Experiments on nine graphs with different degrees of homophily have shown a classification accuracy generally in line with most fully trained models, with extraordinary improvements over two low homophily tasks.
Furthermore, contrary to the theory and experiments that demonstrated the crucial role of system stability in applying GESNs to graph-level tasks, our experiments have shown that node embeddings computed in regions well beyond the theoretical stability threshold are better suited to represent the graph structure.
Future work will analyze more in-depth the embedding space structure, the role of reservoir radius in conditioning the filtering properties of GESN, and the impact of reservoir spectrum.


\begin{footnotesize}


\bibliographystyle{unsrtnat}
\bibliography{bibliography}

\end{footnotesize}


\end{document}